\pdfoutput=1

\documentclass[11pt]{article}

\usepackage[]{acl}

\usepackage{times}
\usepackage{adjustbox}
\usepackage{multirow}
\usepackage{latexsym}
\usepackage{amsmath}
\usepackage{amssymb}
\usepackage{makecell}
\usepackage{algorithm}
\usepackage{algpseudocode}

\usepackage[T1]{fontenc}

\usepackage[utf8]{inputenc}

\usepackage{microtype}
\graphicspath{{./images/}} 
\title{BERT-ASC: Auxiliary-Sentence Construction for Implicit Aspect Learning in Sentiment Analysis}

     \author{Murtadha Ahmed \quad 
	     	Wenbo  \quad 
	     	Shengfeng Pan \quad 
	     	Jianlin Su \quad 
	     	Luo Ao\quad 
	     	Yunfeng Liu\\ 
	     	Zhuiyi AI Lab, Shenzhen, Guangdong, China \\
	     	{\tt \{a.murtadha,brucewen,nickpan,bojonesu,luoao,glenliu\}@wezhuiyi.com}
	     }

\begin{document}
\maketitle
\begin{abstract}
Aspect-based sentiment analysis (ABSA) aims to associate a text with a set of aspects and infer their respective sentimental polarities. State-of-the-art approaches are built on fine-tuning pre-trained language models, focusing on learning aspect-specific representations from the corpus. However, aspects are often expressed implicitly, making implicit mapping challenging without sufficient labeled examples, which may be scarce in real-world scenarios. This paper proposes a unified framework to address aspect categorization and aspect-based sentiment subtasks. We introduce a mechanism to construct an auxiliary-sentence for the implicit aspect using the corpus's semantic information. We then encourage BERT to learn aspect-specific representation in response to this auxiliary-sentence, not the aspect itself. We evaluate our approach on real benchmark datasets for both ABSA and Targeted-ABSA tasks. Our experiments show that it consistently achieves state-of-the-art performance in aspect categorization and aspect-based sentiment across all datasets, with considerable improvement margins. The BERT-ASC code is available at \href{https://github.com/amurtadha/BERT-ASC}{https://github.com/amurtadha/BERT-ASC}.
\end{abstract}

\section{Introduction}\label{sec:intro}

The information provided by individuals on the Web is usually considered {more} trustworthy than that provided by {vendors} \cite{bickart2001internet}. 
{ Aspect-based sentiment analysis (ABSA) entails identifying and analyzing sentiment expressed towards specific aspects or features within text, such as products or services. This approach is essential in applications like customer reviews analysis, where it allows for a nuanced understanding of sentiment towards individual aspects (e.g., service quality, price), enabling businesses to assess overall satisfaction and areas for improvement effectively \cite{pontiki2014semeval,saeidi-etal-2016-sentihood}. There are two types of aspects: aspect-terms, which are explicitly mentioned in the sentence, and aspect-categories, which are rarely explicit and are inferred through indicators \cite{WU2021107073}. Aspect-terms are specific words or phrases directly denoting the object being reviewed, while aspect-categories are broader themes inferred from context and opinion words \cite{pontiki2014semeval}. For example, in the sentence \(s_1\) from Table \ref{tab:running_example}, `coffee' is an aspect-term expressing a positive opinion towards the implicit aspect-category `Food' through the word `outstanding'. Identifying aspect-terms involves locating specific mentions, whereas identifying aspect-categories requires understanding implicit cues in the text.}
Note that the statistics of benchmark datasets, Table \ref{tab:impliti_perc}, show that the majority of sentences express opinions in response to various aspects implicitly (e.g., {using}  the term {`coffee'} to evaluate {`food')}. Therefore, the effective addressing of these sentences would largely determine the performance of ABSA.

The implicit aspect is still a challenging NLP task {in} practice, but it has not received sufficient attention from the research community. The earlier solutions were traditional machine learning-based classifiers {such as}  SVM \cite{kiritchenko2014nrc,brun2014xrce}, which employed feature extraction based on various types of syntactic information{ such as} parser, n-grams, and the sentiment lexicon. However, the aspect category is mostly mentioned implicitly in the text and thus {makes} feature extraction unattainable. 
The state-of-the-art solutions have been built upon various Deep Neural Networks {(DNNs)}. The traditional DNN-based models \cite{ma2018targeted,liu2018recurrent} commonly attempted to learn the aspect-specific representation through various mechanisms (e.g., attention and deep memory). Recently, a considerable shift towards fine-tuning pre-trained language models \cite{sun2019utilizing,wu2020context} has been experienced to jointly address aspect categorization and aspect-based sentiment subtasks. Specifically, ABSA is reformulated into a question-answering task as follows: The text is considered as sentence A in {the} setting of the original BERT \cite{devlin2018bert}, while a query (e.g., { ``Does this sentence describe an opinion towards food?'')} is regarded as sentence B. 
Despite {the} impressive performance of this approach on the aspect-term that appears explicitly in sentence A \cite{karimi2020improving}, the implicit aspect requires mapping each aspect to its indicators and thus relies on sufficient labeled examples, which {may not} be readily available in {real-world }scenarios.

\begin{table*}
	\centering

	\adjustbox{width=\linewidth}{
		\begin{tabular}{l  l lll}
			\hline
			$s_i$& \makecell[c]{Text}&\makecell[l]{Aspect Categories} &BERT-NLI&BERT-ACS \\
			\hline	
			
			$s_1$& Did I mention that the \textbf{coffee} is outstanding?& \{(Food, Pos)\}&\{F\} Food&\{F\} coffee\\
			
			$s_2$&\textbf{Waiters} are very \textbf{friendly} and the \textbf{pasta} is out of this world.& \{(Service, Pos), (Food, Pos)\}&\{F\} service &\{F\} Waiters friendly\\
			$s_3$&I hear that under LOC1 is quite \textbf{cheap} & \{(LOC1, Price, Pos)\}&\{F\} price in LOC1&\{F\} cheap in LOC1\\
			\hline
		\end{tabular}
	}
		\caption{{ Running example with sentences \( s_1 \) and \( s_2 \) from SemEval and \( s_3 \) from SentiHood. For instance, \( s_2 \) expresses positive opinions on Service (waiters) and Food (pasta), while \( s_3 \) describes a positive opinion on the Price of LOC1 (cheap). Words in bold indicate aspect candidates. Comparison of auxiliary sentences for BERT-NLI and BERT-ASC. \{F\} represents "What is the sentiment of ".}}
	\label{tab:running_example}
	
\end{table*}

In this paper, we propose a novel solution to jointly address aspect categorization and aspect-based sentiment subtasks in a unified framework, namely {the} BERT-based Auxiliary Sentence Constructing (BERT-ASC). First, we leverage {Labeled Latent Dirichlet Allocation (L-LDA) }to associate each aspect with a set of seed words in a given corpus. 
Intuitively, the pre-trained language models (PLMs) \cite{devlin2018bert} are trained to map features that often occur in the same context into close points in the embedding space \cite{ahmed-etal-2021-dnn}.
For instance, \textit{``meal"} and \textit{``bread"}, which describe food, are located very close in the latent space. Inspired by this intuition, we exploit the semantic distribution of the seeds in the embedding space to capture more coherent indicators. Specifically, we construct {an} auxiliary sentence that relies on the semantic distance between the aspect's seed and the sentence on-target. 
{Finally, we fine-tune the PLM (e.g., BERT) to learn aspect-specific representations based on automatically constructed auxiliary sentences, rather than the aspect itself (as adopted in \cite{sun2019utilizing}). Given \( s_1 \) in Table \ref{tab:running_example}, BERT-NLI trains with the auxiliary sentence "What is the sentiment of food?", while our BERT-ASC uses the generated auxiliary sentence "What is the sentiment of coffee?". This approach simplifies the mapping process between indicators and their implicit aspects, thereby reducing the learning cost.}
\begin{table}
	\centering

	\begin{tabular}{l c c}
		\hline
		\multirow{2}{*}{Dataset}& \multicolumn{2}{c}{ Implicit Aspect Percentage }\\ \cline{2-3}
		&Train& Test\\
		\hline
		SemEval&77.6\%&73.8\%\\
		SentiHood&96.8\%&97.5\%\\
		
		\hline	
	\end{tabular}
		\caption{{The percentage of sentences that express opinions towards various aspects implicitly in the SemEval and Sentihood datasets is notable. A category is considered an implicit aspect if it is not explicitly mentioned in the text. For example, in the sentence \textit{"the sushi is yummy"}, the aspect { of}  food is implied but not directly stated.}}
	\label{tab:impliti_perc}
	
\end{table}

In brief, the main contributions are three-fold: 
\begin{enumerate}
	
	\item We present a simple but effective mechanism to construct {an} auxiliary sentence for the implicit aspect based on its seed semantic distribution in the embedding space;
	\item We introduce a BERT-based fine-tuning approach that jointly addresses aspect categorization and aspect-based sentiment {color{red} analysis} in a unified framework. The model is trained to learn the aspect-specific representation in response to the automatically constructed auxiliary sentence, instead of the aspect itself.
	\item We evaluate the proposed solution on benchmark datasets for TABSA and ABSA tasks. Our extensive experiments show that it consistently achieves state-of-the-art performance in terms of aspect categorization and aspect-based sentiment {analysis} subtasks across all test datasets, and the improvement margins are considerable.
	
\end{enumerate}

The remainder of the paper is organized as follows. Section \ref{sec:related_work} reviews related work. Section \ref{sec:approach} describes {the} proposed solution. Section \ref{sec:experiments} presents the experimental settings and {empirically} evaluates the performance of the proposed solution. Finally, we conclude this paper with Section \ref{sec:conclusion}.

\section{Related work}\label{sec:related_work}

Aspect-based sentiment analysis is a fine-grained classification problem \cite{pontiki2014semeval,saeidi-etal-2016-sentihood}. Unlike the explicit aspect-term, the aspect-category is mostly described implicitly, {making}  learning its representation a more challenging task. {The} implicit aspect is a challenging NLP task {in} practice but has not received sufficient attention from the research community. { It is noteworthy that there exists a closely related task known as Target-Aspect-Sentiment Detection (TASD), which focuses on identifying target-aspect-sentiment triples within a given sentence. Previous approaches to TASD, such as those presented in \cite{WU2021107073} and \cite{wan2020target}, have typically divided the problem into two distinct subtasks: binary text classification and sequence labeling. These methods employed a single neural model built upon BERT, aiming to minimize a combined loss function that addresses both subproblems.
	Recently, a novel task has emerged in the field, termed Aspect-Category-Opinion-Sentiment (ACOS) Quadruple Extraction, as introduced in \cite{cai-etal-2021-aspect}. This task is supported by two datasets: Restaurant-ACOS and Laptop-ACOS. The primary objective of ACOS Quadruple Extraction is to identify and extract all aspect-category-opinion-sentiment quadruples present within a review sentence. This task holds significant value for Aspect-Based Sentiment Analysis (ABSA) as it provides insights into implicit aspects and their associated opinions.}

The unsupervised techniques commonly {tackle}  the task as sequence labeling based on prior knowledge (e.g., WordNet and {manually} constructed category's seeds). Co-occurrence association rule mining approaches \cite{schouten2017supervised} attempt to enable activation value spreading between tokens in the same sentence. However, these approaches heavily depend on handcrafted features and, unfortunately, some categories, e.g., {`general'} and {`miscellaneous'}, are very abstract. The supervised approaches address the task as a multi-class classification problem. SVM-based models \cite{castellucci2014unitor,kiritchenko2014nrc} introduce a set of features, including n-grams, syntax information, and lexicon features. However, these techniques cannot {accurately} capture the semantic context of different aspects

Last decade, various approaches have been proposed to learn aspect-specific representations in a corpus. A hybrid method \cite{zhou2015representation} introduced modeling semantic relations based on domain-specific embeddings as hybrid features for a logistic regression classifier. A Convolutional Neural Network-based features model \cite{toh2016nlangp} incorporated automatically learnable features with n-grams and POS tags to train one-vs-all linear classifiers. An LSTM equipped with a CNN layer model \cite{xue2017mtna} addressed aspect categorization and sentiment subtasks simultaneously. A Gated Recurrent Units model equipped with a topic attention mechanism \cite{movahedi2019aspect} proposed filtering aspect-irrelevant information away. A BERT-based fine-tuning model \cite{sun2019utilizing} was proposed to address the task as question answering. A context-guided BERT-based fine-tuning approach \cite{wu2020context}, which adopted a context-aware self-attention network, was introduced to learn {distributing}  attention under different contexts. 
{
	\cite{VenugopalanG22} introduced an unsupervised approach for aspect term extraction using guided LDA with manually aspect seed words and enhanced by linguistic rule-based regular expressions. Graph Neural Networks (GCNs) \cite{li2021dual,dai2021does,ZhaoTTWC22} were introduced to integrate syntax information into the learning process to enhance aspect-context capturing. Incorporating implicit sentiment and explicit syntax knowledge via self-supervised pre-training \cite{LiZJLSTT22}. A domain adaptation method that retrieves and edits prototypes from unlabeled target data to enhance word transferability, enabling cross-domain aspect term extraction and sentiment classification \cite{ChenQ22}.    A hierarchical dual GCNs model with two GCNs for extracting syntactic and semantic information. It uses a self-attention matrix for semantic extraction and a dependency feature-aware GCN for syntactic mapping, with multiple layers to capture various linguistic features \cite{Zhou0CC23}. }
However, these approaches, at least in their current settings, struggle with implicit aspect detection. While they have shown impressive performance in identifying explicitly mentioned aspect-terms within sentences, handling implicit aspects remains challenging. Detecting implicit aspects requires mapping each aspect to its indicators, a process that depends heavily on having a sufficient number of labeled examples. Unfortunately, such labeled examples are often scarce in real-world scenarios \cite{ahmed-etal-2021-dnn}.
{

	In-Context Learning (ICL) allows Large Language Models (LLMs) to perform tasks efficiently by using in-context examples without fine-tuning \cite{BrownMRSKDNSSAA20,HanZDGLHQYZZHHJ21,Qiu-2003-08271}. This approach has influenced sentiment analysis through dense demonstration retrieval and context-based extension. Dense retrieval uses vectors for semantic matching to enhance LLM performance by retrieving relevant examples \cite{reimers2019sentence, wang2022simlm}. 
	LLM-R introduced iterative training with a reward model and knowledge distillation for a bi-encoder-based dense retriever. However, task-specific retrievers may require lengthy training. Recent advancements like structured prompting by \citet{Yaru-2212-06713} and novel position embeddings by \citet{pcw} aim to improve integration of examples into LLMs. \citet{nbce} proposed context splitting and a voting mechanism for relevant context selection. However, handling aspect detection and aspect sentiment simultaneously remains challenging, as both tasks typically rely on the same input data and are interdependent.

}

To overcome this limitation, BERT-ASC introduces the concept of constructing an auxiliary sentence, which encourages the model to encode sentences in response to generated auxiliary sentences. This strategy enhances the model's ability to capture aspect-specific representations, even when the aspects are not explicitly mentioned in the text. { Most similar to our work in incorporating auxiliary sentences is BERT-NLI \cite{sun2019utilizing}. However, in BERT-NLI, the auxiliary sentence is constructed by questioning about the category on target. For example, consider sentence \(s_1\) "Did I mention that the coffee is outstanding?" from the running example in Table \ref{tab:running_example}; their auxiliary sentence might be, "What do you think of the food?". We believe this approach is more suitable for aspect-terms where the aspect is explicitly expressed. In contrast, we first extract representative words for the aspect category, such as 'coffee' for 'food' in \(s_1\), and then construct the auxiliary sentence as: "What do you think of the coffee?". This approach alleviates the need for BERT to map the category to its related terms, thereby potentially boosting performance.
}

\section{The Proposed Solution}\label{sec:approach}
\begin{figure*}
	\centering
	\includegraphics[scale=.65]{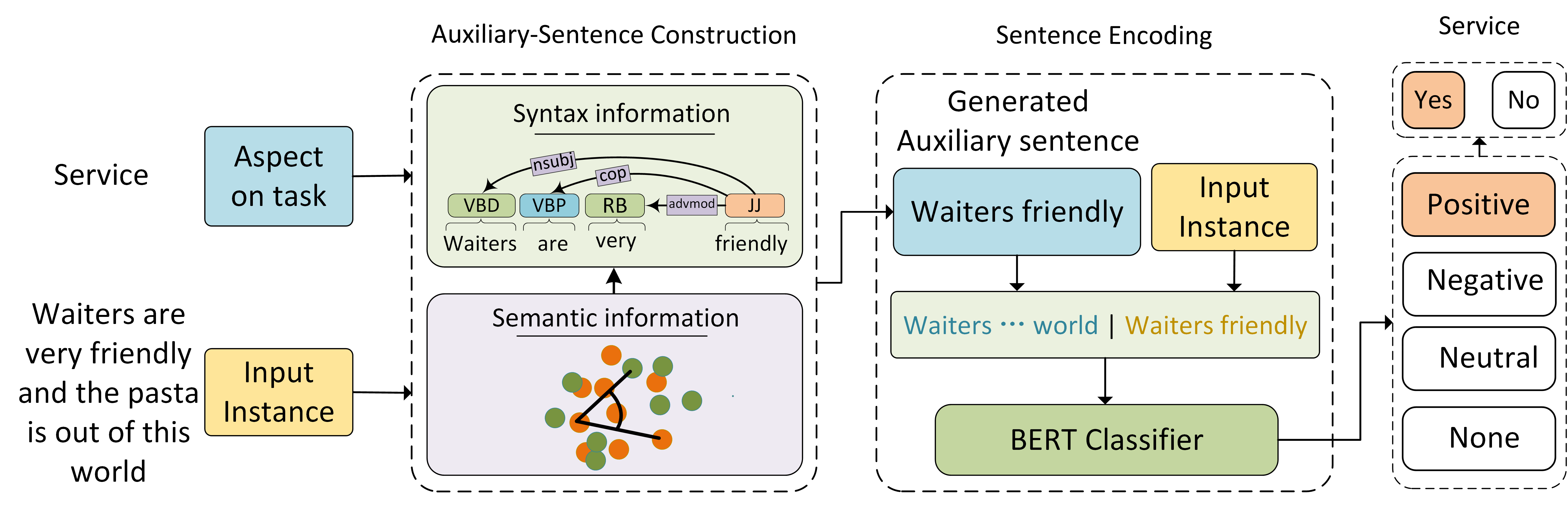}	
	
	\caption{{ illustrative example of BERT-ASC. It consists of two steps: constructing an auxiliary sentence and jointly training for aspect and sentiment classification. Consider sentence \( s_2 \) from the running example in response to "service." The generated auxiliary sentence includes the semantic candidate \textit{waiters} and its syntactic modifier \textit{friendly}. The aspect categorization layer processes the output of the aspect sentiment layer and applies binary classification to determine whether the discussed aspect is on-target in the input sentence. The label \textit{none} denotes \( 0 \), while any other label denotes \( 1 \).}}
	
	\label{fig:framework}
\end{figure*}

We begin by defining both TABSA and ABSA tasks, then present the technical details of the proposed solution shown in Figure \ref{fig:framework}. {To provide a clearer understanding of the proposed approach, the workload is illustrated in pseudo-code format in Algo \ref{algo:workload}.}  {In particular,}  it first leverages supervised LDA to extract a set of seeds for each aspect. Then, it generates an auxiliary sentence for the aspect on-target by modeling the semantic relations to its seed. Note that auxiliary sentence construction is carried out off-line and before the training process begins. Finally, it fine-tunes BERT to learn the aspect-specific representation in response to the automatically {constructed}  auxiliary sentence. We will now proceed to describe this process in greater detail.
{
	\subsection{Task Definition}
	\begin{enumerate}
		\item \textbf{TABSA Task.}
		We formulate the SentiHood dataset \cite{saeidi2016sentihood} as a TABSA task. Given a sentence review $s$ that consists of a sequence of words $\{w_1, w_2, ..., w_n\}$ in which $m$ words $\{w_i, ..., w_m\}$ are from a set {$T$}  of $k$ pre-identified targets $\{t_1, ..., t_k\}$, the goal is to predict the sentiment for each aspect category associated with each unique target explicitly mentioned in the sentence.
		Given a sentence review $s$ and a predefined target list {$T$} and a list of aspect categories  \textit{$C =\{$price,transit-location,safety,general$\}$}, the model  is required to label each pair  $\{(t,c) : (t \in {T}, c \in C)\}$ with \textit{$y\in \{$negative, positive, none$\}$}. Note that the model predicts a single sentiment label for each unique target-aspect pair in a sentence.
		
		\item \textbf{ABSA Task.}
		We formulate the SemEval-2014 Task 4 dataset \cite{pontiki2014semeval} as {an} ABSA task. The target-aspect pairs $\{t, c\}$ of TABSA { become}  only an aspect category $c$. Given a review sentence $s$, the model attempts to predict a sentiment label \textit{$y \in \{$negative, neutral, positive, conflict, none$\}$} for each aspect category $\{c:(c \in C)\}$ with a predefined category list \textit{C=$\{$food, price, service, ambience, anecdotes$\}$}.
	\end{enumerate}
	
}

\subsection{Aspect Seed Extraction}
\begin{table}
	\centering

		\adjustbox{width=\linewidth}{
		\begin{tabular}{ll}
			\hline
			Aspect& \makecell[c]{Seed}\\
			\hline
			Food& delicious, chicken, menu, beef, sushi\\
			Price&charge, cheap, reasonable, bill, inexpensive\\
			Service&waiters, attentive, rude, reservations, staff\\
			Ambience&crowded, decor, loud, atmosphere, scene\\
			\hline
		\end{tabular}
			}
		\caption{An example of top-5 seed for SemEval 2014 task 4 dataset.}
	\label{tab:seeds}
\end{table}

A significant proportion of the text is tagged with different aspects. In other words, the implicit aspect exists in the corpus through a set of explicit indicators, also referred to as seeds. Previous work manually built a list of seeds for each category, which was expanded based on prior knowledge (e.g., WordNet) \cite{Schouten2016tkde,ahmed2020constructing}. To minimize additional human effort, we leverage the labeled corpora to extract the aspect's seeds. For this purpose, we adopt Labeled LDA (L-LDA) \cite{ramage-etal-2009-labeled}. Specifically, for each document $d \in D$, traditional LDA typically assigns a multinomial mixture distribution $\theta^{(d)}$ over all $K$ topics from a Dirichlet prior $\alpha$. However, L-LDA uses the labeled corpus to constrain $\theta^{(d)}$ to be defined only over the topics (i.e., the aspects in this context) that correspond to its labels ${\Lambda^{(d)}}$. The technical details are well-explained in \cite{ramage-etal-2009-labeled}. An example of the top-5 extracted seeds for the SemEval 2014 Task 4 dataset is illustrated in Table \ref{tab:seeds}.

\subsection{Auxiliary-Sentence Construction}\label{aspect_related}

\begin{figure}
	\centering
	\includegraphics[scale=.6]{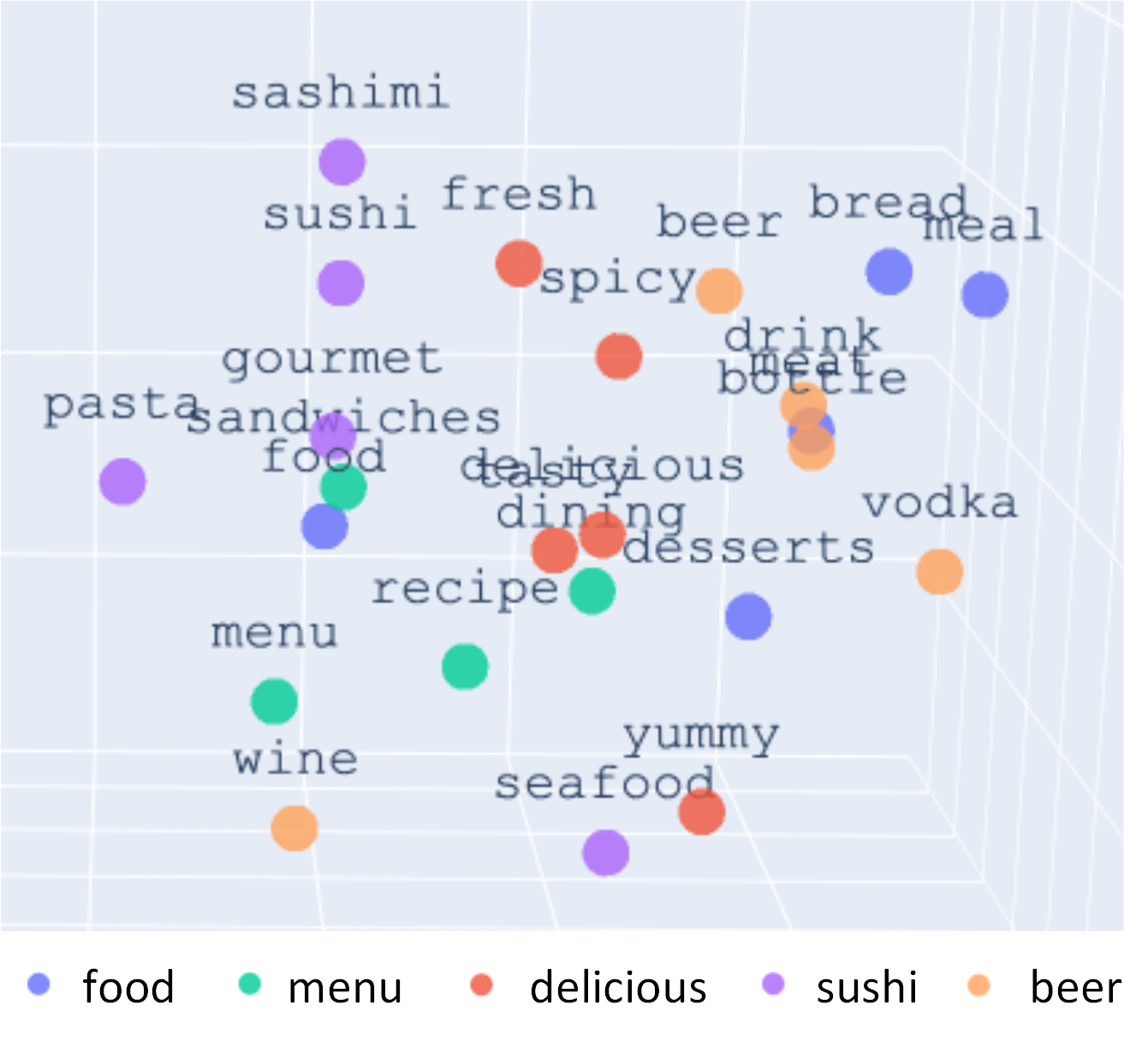}	
	\caption{An illustrative example of the {food's}  coherent distribution in the embedding space.}
	\label{fig:food_seed}
\end{figure}

As the ultimate goal is to jointly address aspect categorization and aspect-based sentiment subtasks simultaneously, we thus incorporate the semantic context and syntax information to construct the auxiliary sentence as shown in Figure \ref{fig:framework}. In other words, aspect categorization relies on the semantic relations between sequences in response to a given aspect, while aspect-based sentiment{ focuses on}  the aspect-opinion words, also called \textit{modifiers} \cite{liu2012sentiment,hu2004mining,li2021dual}. Consider the sentence $s_1$ of the running example {in}  Table \ref{tab:running_example}, Food is semantically indicated through the semantic term coffee, while the opinion-word outstanding sentimentally expresses its polarity. To this end, we first capture the semantic candidates, then syntactically generate their opinion-words as follows.

\subsubsection{Semantic candidates} 
Language models \cite{mikolov2013linguistic,devlin2018bert} are trained to map the features that often occur in the same context into close points in the latent space \cite{he2017unsupervised,ahmed2020constructing}. Figure \ref{fig:food_seed} depicts the distribution of food-opinion words in the latent space.
As can be seen, the words are located very closely, which enables {us}  to leverage the semantic distance.
Therefore, we apply the semantic distance between the seed of the aspect on-target and the input sentence. Then, the words that meet a pre-specified threshold (e.g., 0.8) are considered as semantic candidates. Note that the threshold can be specified using the validation set. Consider sentence $s_1$ of the running example in Table \ref{tab:running_example}, given the aspect \textit{food}, the term \textit{coffee} is semantically very close to the seed \textit{menu} in the embedding space. Thus, we consider \textit{coffee} as a semantic candidate for the aspect \textit{food}. Next, we exploit the syntactic information to extract the opinion words.

\subsubsection{Syntactic information}
\begin{figure}
	\centering
	\includegraphics[scale=.4]{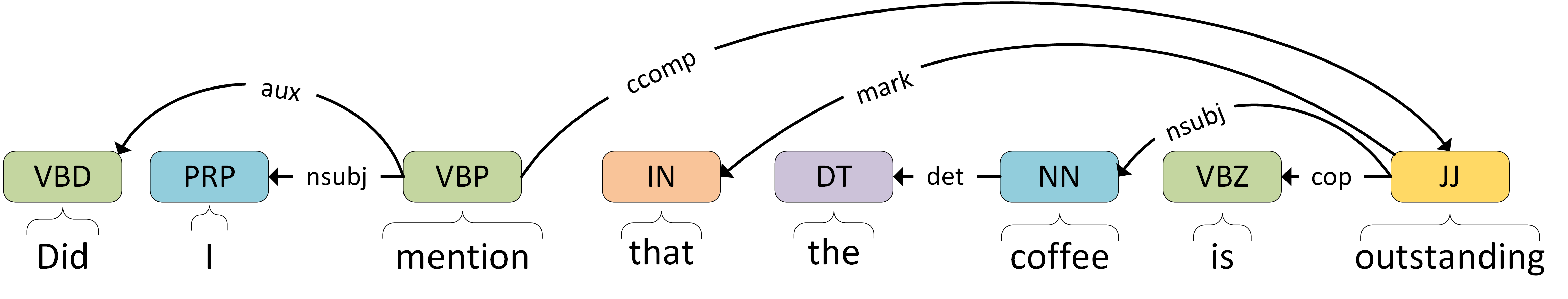}	
	\caption{{An example of the dependency parser for sentence $s_1$ from the running example. POS abbreviations used include: VBD (past tense of a verb), PRP (pronoun), VBP (present tense, non-third person singular form of a verb), IN (preposition), DT (determiner), NN (noun), VBZ (present tense, third person singular form of a verb), JJ (adjective). The dependency links represented are: nsubj (subject of a verb), aux (auxiliary verb), ccomp (complement clause), det (determiner), mark (clause marker), cop (copula verb mark).}}

	\label{fig:dep_tree}
\end{figure}

Now that we obtain the semantic candidates, we leverage the dependency tree to extract target-opinion words, which has been widely {adapted}  in previous work \cite{hu2004mining,popescu2007extracting,bloom2007extracting,qiu2011opinion,zeng2019variational}. Figure \ref{fig:dep_tree} depicts the dependency relations of sentence $s_1$ in the running example.
We designed three rules to capture the syntactic information of the semantic candidates as follows:

\begin{itemize}

	\item We consider the opinions that share {the}  grammatical relation \textit{amod} (adjectival modifier) \cite{de-marneffe-manning-2008-stanford}. For example, in the phrase ``very delicious sushi'', we extract \textit{delicious} as a syntactic modifier.
	\item We include the opinions satisfying the relation \textit{nsubj} (nominal subject). Note that \textit{nsubj} may appear between various POS (e.g., nouns, adverbs, or adjectives), we only include adjectives that modify nouns. For example, in the sentence ``The sushi is yummy,'' we can extract \textit{yummy} as a modifier.
	\item We also include the outlier modifiers that satisfy the grammatical relation \textit{advmod} (adverb modifier). For example, ``Genetically modified food,'' we extract \textit{Genetically} as an outlier modifier of the opinion word \textit{modified}.

\end{itemize}

In brief, the automatically constructed auxiliary sentence given the aspect \textit{food} is \textit{``coffee is outstanding''}. It is noteworthy that the auxiliary-sentence is initialized to the aspect itself in case no candidate is detected. {Finally}, we encourage the BERT model to learn implicitly the aspect-specific representation in response to the constructed auxiliary-sentence instead of the aspect itself, as in \cite{sun2019utilizing}.

\subsection{BERT-based classifier}

Recently, BERT\footnote{\url{https://github.com/google-research/BERT}} \cite{devlin2018bert} achieved state-of-the-art performance in the ABSA task and has been widely utilized in many NLP tasks \cite{davison2019commonsense,peters2019tune,ahmed2023rank}. As BERT was trained on a large corpus to learn the semantic context, a fine-tuning phase is thus needed to adapt the modeling (i.e., understanding) to a specific domain \cite{xu2019BERT,du2020adversarial}. Our goal is to learn the representation in response to a given aspect, i.e., the query in BERT's notation \cite{devlin2018bert}.
In our implementation, we concatenate the automatically generated auxiliary-sentence \(a\) of the aspect on-target with the sentence review \(s\).

Let \[
x = (\text{[CLS]}, a_1, \dots, a_m, \text{[SEP]}, s_1, \dots, s_n, \text{[SEP]})
\]where \(a_1,...,a_m\) is the auxiliary-sentence (with \(m\) tokens) and \(s_1,...,s_n\) is a review sentence, \text{[CLS]} and \text{[SEP]} are special tokens. We feed \(x\) to BERT:
\begin{equation}\label{eq:BERT}
	Z = \text{BERT}(x),
\end{equation}
where \(Z\) denotes the hidden layers. BERT-based models mostly consider the \(h_{\text{CLS}}\) token as the sentence representation. However, the authors of \cite{jawahar-etal-2019-BERT} have investigated the language structure learned by BERT and have found that semantic features can be extracted from the top layers. Inspired by Feature Pyramid Networks (FPNs) \cite{lin2017feature}, we follow \cite{karimi2020improving} by exploiting the top four layers of BERT. Specifically, one more BERT layer is added that takes both the previous and the current layers as input and performs predictions for each layer separately. The intuition behind this architecture is that the deeper layers contain more semantic information in response to the category on-target \cite{lin2017feature}. Note that BERT-ASC requires \(\vert \mathcal{C} \vert\) (i.e., the number of aspects) forward passes during inference time, while standard fine-tuning only requires one forward pass to get the prediction for all aspects. In fact, the inference speed can be easily improved by reusing the siamese transformer networks \cite{reimers2019sentence}.

\begin{algorithm}[t]
	\caption{Training Procedure}
	\begin{algorithmic}[1]
		
		\State \textbf{Input:} Corpus $C$, Aspects $A$, Pre-trained BERT model
		\State \textbf{Output:} Fine-tuned BERT model for aspect-specific representation
		
		\Procedure{ExtractSeeds}{$C$, $A$}
		\For{each aspect $a \in A$}
		\State Extract set of seeds for $a$ using L-LDA
		\EndFor
		\State \textbf{Return} Seed sets for each aspect
		\EndProcedure
		
		\Procedure{ConstructAuxiliarySentences}{Seed sets}
		\For{each aspect $a \in A$}
		\State Generate auxiliary sentence for $a$
		\EndFor
		\State \textbf{Return} Auxiliary sentences for each aspect
		\EndProcedure
		
		\Procedure{FineTuneBERT}{}
		\State Load pre-trained BERT model
		\For{each aspect $a \in A$}
		\State Retrieve auxiliary sentence for $a$
		\State Fine-tune BERT using auxiliary sentence for $a$
		\EndFor
		\State \textbf{Return} Fine-tuned BERT model
		\EndProcedure
		
	\end{algorithmic}
	\label{algo:workload}
\end{algorithm}

{We present the training procedure in Algo \ref{algo:workload}.  Time Complexity: LDA typically has a complexity of \( O(KDN) \), where \( K \) represents the number of topics (or aspects), \( D \) is the count of documents, and \( N \) signifies the average number of words in each document. Adding a labeling step in supervised LDA does not substantially alter the overall complexity order. For the construction of auxiliary sentences, the time complexity is mainly influenced by the number of words in the input sentence and the number of seed words associated with the aspect. Assuming there are \( W \) words in the sentence and \( S \) seed words, and given that the calculation of semantic distance is constant in time, this phase would have a time complexity of \( O(WS) \). Space Complexity: The primary space demands are attributed to storing the document-word matrix and the topic-word distributions, generally amounting to \( O(DN + KV) \), where \( V \) is the size of the vocabulary. In the case of supervised LDA, additional space is needed for label-topic distributions; however, this typically constitutes a marginal increase compared to the document-word matrix's size. The space required for constructing auxiliary sentences mainly involves storing the seed words and the words in the input sentence, which is \( O(W + S) \).}
\section{Empirical  Evaluation}\label{sec:experiments}

\subsection{Datasets}

\begin{table}
	\centering

		\adjustbox{width=\linewidth}{
		\begin{tabular}{lcccc}
			\hline
			Dataset&Train &Test &\# Aspects&  MA Prop (\%) \\
			\hline
			SemEval &3,041&800&5&23.65 \% \\
			SentiHood &3,724&1,491&4&31.0 \%\\
			\hline
		\end{tabular}	
		}
		\caption{The statistics of datasets (sentence).  MA Prob indicates the percentage of sentences labeled with more than one category.}
	\label{tab:datasets}
\end{table}

We evaluate our solution with benchmark datasets in English for both TABSA and ABSA tasks. For the TABSA task, we used the SentiHood dataset\footnote{SentiHood: \url{https://github.com/uclnlp/jack/tree/master}.} \cite{saeidi2016sentihood}, which was built from the { Question Answering Yahoo corpus with location names from London and the UK}. The dataset consists of sentence reviews that evaluate at least one aspect category \(c\) in response to the target \(t\). For each sentence, our solution predicts the label \(y\) for each target-aspect pair \((t, c)\). For the ABSA task, we used the benchmark dataset introduced by SemEval 2014 Task 4\footnote{SemEval-2014 Task 4: \url{https://alt.qcri.org/semeval2014/task4/}.} \cite{pontiki2014semeval}, which consists of restaurant reviews collected from the { Citysearch New York corpus}. Each sentence review is annotated with a set of aspect categories and their respective polarities. { Each dataset is partitioned into train, validation, and test sets} as described in its original paper. The statistics of the datasets are shown in Table \ref{tab:datasets}.

\subsection{Experimental Settings}

{ We use BERT as our baseline, which has been widely employed by  the comparative baselines. It is important to note that other language models, such as RoBERTa \cite{abs-1907-11692} or Sentence-BERT \cite{reimers2019sentence}, can be utilized in a similar manner for this purpose.}
We follow the settings of the original BERT-base model \cite{devlin2018bert}; { our model consists of 12 heads and 12 layers with a hidden layer size of 768,} and the total number of parameters for our model is 138M. When fine-tuning, we keep the dropout probability at 0.1 and set the number of epochs to 3. The initial learning rate is \(3e^{-5}\) for all layers with a batch size of 32.
For the seed extraction, we used the open-source code of L-LDA \footnote{L-LDA: \url{https://github.com/JoeZJH/Labeled-LDA-Python}} \cite{ramage2009labeled} and select the top-10 representatives for each aspect. We used the Stanford CoreNLP as the dependency parser. For the auxiliary sentence construction, we initialize the feature vectors with word vectors trained by word2vec with negative sampling on each dataset, setting the embedding size to 200, window size to 10, and negative sample size to 5. { The semantic distance threshold is set to a value of 0.3 and 0.4} for the SemEval and SentiHood datasets, respectively. Note that these values are set based on the validation set.

We follow the experimental settings of the tasks' definition \cite{pontiki2014semeval,saeidi2016sentihood}, which were widely adapted in the comparative methods \cite{sun2019utilizing,wu2020context}. We define two subtasks for each dataset as follows: The aspect categorization subtask aims to detect the aspect described in a given sentence. Note that the label \textit{none} represents the absence of a given aspect in the sentence. The aspect-based sentiment classification subtask aims to associate each detected aspect { (i.e., ignore the \textit{none} label)} with its respective polarity { (i.e., \textit{negative} and \textit{positive} for Sentihood, and \textit{negative, neutral, positive, conflicting} for SemEval)}.

\subsection{Exp-I: ABSA}
During the evaluation with the ABSA task, we follow the task's description \cite{pontiki2014semeval} and the comparative methods \cite{sun2019utilizing,wu2020context}. For aspect categorization, we report Precision, Recall, and F1 scores.
For aspect-based sentiment classification, we report the accuracy metrics for three different evaluations: binary classification { (i.e., negative or positive)}, 3-class classification { (i.e., negative, neutral, or positive)}, and 4-class classification { (i.e., negative, neutral, positive, or conflict)}. We compare our solution to the systems that achieved the best performance in the SemEval competition and the BERT-based state-of-the-art solutions:

\begin{table*}[t]
	\centering

	\adjustbox{width=\linewidth}{
		\begin{tabular}{lcccccccc}
			\hline
			\makecell[c]{\multirow{2}{*}{Model}}&\multicolumn{3}{c}{Aspect Categorization}&& \multicolumn{3}{c}{Aspect Sentiment}\\\cline{2-4}\cline{6-8}
			&Precision&Recall&F1 score&&Binary&3-Class&4-Class\\
			\hline
			XRCE \cite{brun2014xrce}&	83.23&	81.37&	82.29&&	-&	-&	78.1\\
			NRC-Canada\cite{kiritchenko2014nrc}&	91.04&	86.24&	88.58&&	-&	-&	82.9\\
			BERT-single \cite{devlin2018bert} &	92.78&	89.07&	90.89&&	93.3&	86.9&	83.7\\
			BERT-NLI-M\cite{sun2019utilizing} &	93.15 &90.24&91.67&&94.4&  88.7&  85.1\\
			CG-BERT \cite{wu2020context}& 93.02&  90.00&  91.49&&  94.3&  89.9&  85.6\\ 
			BERT-NLI-B \cite{sun2019utilizing}&93.57& 90.83& 92.18&&	95.6&	89.9&	85.9\\ 
			QACG-BERT \cite{wu2020context}&	94.38&	90.97&	92.64&&	95.6&	90.1&	86.8\\
			
			BERT-ASC					&\textbf{95.56$\pm$1.07}	&\textbf{92.96$\pm$1.71}	&\textbf{94.22$\pm$0.49}&&\textbf{95.83$\pm$0.49}&\textbf{91.21$\pm$0.46}&\textbf{87.44$\pm$0.30}\\

			\hline
			
		\end{tabular}
	}
		\caption{Model performance on the SemEval-2014 Task 4 ABSA dataset, with { the} best performances { highlighted} in bold. Aspect categorization and sentiment classification correspond to Subtask 3 and Subtask 4, respectively. { The} mean and standard deviation of 5 runs with different { seeds are} reported. The comparative results are retrieved from the original papers.}
	\label{tab:ABSA_Results}
	
\end{table*}

\begin{table*}[t]
	\centering

	\adjustbox{width=\linewidth}{
		\begin{tabular}{llccccccc}
			\hline
			\makecell[c]{\multirow{2}{*}{PLM}}&\makecell[c]{\multirow{2}{*}{Model}}&\multicolumn{3}{c}{Aspect Categorization}&& \multicolumn{3}{c}{Aspect Sentiment}\\\cline{3-5}\cline{7-9}
			&&Precision&Recall&F1 score&&Binary&3-Class&4-Class\\
			\hline
			\multirow{2}{*}{\(\text{BART}_{\text{Base}}\)}
			&BERT-NLI&95.20$\pm$0.58&81.09$\pm$3.41&87.53$\pm$1.76&&95.62$\pm$0.47&90.90$\pm$0.49&86.90$\pm$0.42\\
			&BERT-ASC&96.36$\pm$0.47&91.68$\pm$0.99&93.96$\pm$0.47&&95.14$\pm$0.46&90.70$\pm$0.33&86.32$\pm$0.13\\
			\hline
			\multirow{2}{*}{\(\text{RoBERTa}_{\text{Base}}\)}
			&BERT-NLI&94.22$\pm$1.36&85.64$\pm$2.92&89.68$\pm$1.08&&95.70$\pm$0.41&91.78$\pm$0.72&88.58$\pm$0.50\\
			&BERT-ASC&96.61$\pm$0.72&92.96$\pm$1.77&94.74$\pm$0.80&&95.56$\pm$0.14&91.55$\pm$0.24&87.78$\pm$0.22\\
			\hline
			\multirow{2}{*}{\(\text{BART}_{\text{Large}}\)}
			&BERT-NLI&95.44$\pm$0.71&85.52$\pm$3.55&90.16$\pm$1.74&&97.10$\pm$0.56&93.19$\pm$0.87&89.83$\pm$0.59\\
			&BERT-ASC&96.61$\pm$0.58&92.74$\pm$2.08&94.61$\pm$0.89&&97.04$\pm$0.08&93.63$\pm$0.24&89.80$\pm$0.26\\
			\hline
			\multirow{2}{*}{\(\text{RoBERTa}_{\text{Large}}\)}
			&BERT-NLI&95.80$\pm$0.40&81.89$\pm$1.27&88.30$\pm$0.58&&97.21$\pm$0.17&93.48$\pm$0.05&89.90$\pm$0.25\\
			&BERT-ASC&96.20$\pm$0.92&93.46$\pm$1.45&94.79$\pm$0.53&&96.68$\pm$0.20&92.70$\pm$0.45&89.02$\pm$0.21\\
			\hline

			\multirow{2}{*}{\(\text{BERT}_{\text{Large}}\)}&BERT-NLI&93.34$\pm$0.92&81.70$\pm$3.51&87.08$\pm$1.76&&95.08$\pm$0.23&90.50$\pm$0.64&86.76$\pm$0.68\\
			&BERT-ASC&95.19$\pm$1.34&94.48$\pm$2.85&94.79$\pm$0.82&&94.48$\pm$0.17&90.36$\pm$0.51&86.20$\pm$0.43\\
			\hline
			
			\hline

		\end{tabular}
	}
		\caption{{Evaluation results of SemEval on various PLMs. The results are averaged over five runs with different seeds, with standard deviation provided. As can be seen, BERT-ASC significantly improves aspect detection and achieves competitive results in sentiment detection. All results are from our implementations.}}
	\label{tab:semeval_plm_acd}
	
\end{table*}
\begin{itemize}

	\item \textbf{XRCE} \cite{brun2014xrce}. An SVM-based approach { that} relied on features extracted from { a} syntactic information parser and BOW. For the aspect categorization subtask, it trained a logistic regression-based model to compute the probability of { belonging} to a given aspect.
	
	\item \textbf{NRC-Canada} \cite{kiritchenko2014nrc}.  NRC-Canada reported the best performance in the SemEval 2014 competition task 4. It is an SVM-based approach that relied on feature extraction, including ngrams, non-contiguous ngrams, and lexicon features{, etc.}
	\item \textbf{BERT-single} \cite{devlin2018bert}. A BERT-based baseline that takes the sentence as input and addresses the task as a multi-label classification problem.
	\item \textbf{BERT-NLI-M} \cite{sun2019utilizing}. A sentence-pair BERT-based fine-tuning model, which was trained to learn the aspect-specific representation as a pseudo-sentence natural language inference.
	\item \textbf{CG-BERT} \cite{wu2020context}. It is a context-guided BERT-based fine-tuning model, which adopted a context-aware self-attention network. It thus introduced { a method} to learn { how to distribute} the attention under different contexts.
	\item \textbf{BERT-NLI-B} \cite{sun2019utilizing}. It is a variant of BERT-pair-NLI that concatenated the aspect category with each sentimental valance as an auxiliary sentence and then addressed the task as a binary classification problem { (i.e., \(y \in \{yes, no\}\))} to obtain the probability distribution.

	\item \textbf{QACG-BERT}  \cite{wu2020context}. QACG-BERT is an improved variant of the CG-BERT model that introduced learning quasi-attention weights  in a compositional manner to enable subtractive attention lacking in softmax-attention.
	
\end{itemize}
\subsubsection{Results}
\begin{table*}
	\centering

		\begin{tabular}{lcccccccc}
			\hline
			\makecell[c]{\multirow{2}{*}{Model}}&\multicolumn{3}{c}{Aspect Categorization}&& \multicolumn{3}{c}{Aspect Sentiment}\\\cline{2-4}\cline{6-8}
			&Precision&Recall&F1 score&&Binary&3-Class&4-Class\\
			\hline
			BERT-ASC W/O Semantic	&	95.03 &90.24&	92.52 &&95.21 &90.16 &	85.92\\
			BERT-ASC W/O Syntactic	&	95.71 &91.41&	93.52 &&95.43 &90.92 &	86.71\\
			BERT-ASC 				&	95.56 &92.96&	94.22 &&95.83 &91.21 &	87.44\\
			\hline
			
		\end{tabular}
		\caption{{Ablation study by eliminating semantic information (BERT-ASC W/O Semantic) and syntactic information (BERT-ASC W/O Syntactic). BERT-ASC W/O Semantic demonstrates a greater contribution to the final performance due to the effective ability of DNNs to model implicit information. However, BERT-ASC W/O Syntactic may leave many examples with opinion words due to the inaccuracy of the dependency parser as shown in Table \ref{tab:error_analysis_1}.}}
	\label{tab:abl}
	
\end{table*}

We use the dev set to select the best model and average 5 run performances with different seeds, and report the detailed evaluation results in Table~\ref{tab:ABSA_Results} from which we have made the following observations. (1) The traditional machine learning-based approaches  XRCE and NRC-Canada report the lowest performance due to the limited ability in context understanding of these models{;} (2) The baseline BERT-single cannot perform as accurately compared to the other BERT-based approaches due to the absence of the auxiliary-sentence, which boosts BERT's ability to capture the context of the aspect{;} (3) For the aspect categorization task, our solution consistently achieves state-of-the-art performance in terms of Precision, Recall, and F1 scores. Besides, our solution gives the best recall score with a { 1.07\% improvement} over the previous state-of-the-art model, and thus demonstrates the ability of the proposed solution to accurately detect the correct category{;} (4) For the aspect-based sentiment task, our solution outperforms the state-of-the-art model in terms of binary and multi-class performance. As can be seen, we report the best performance on 3-class (i.e., negative, neutral, or positive), which are the most dominant labels in sentiment analysis with a { 1.6\% improvement}. Due to the widely recognized challenge of sentiment analysis, the achieved improvements can be deemed very considerable. The experimental results suggest that a well-designed auxiliary-sentence can boost BERT's performance on the ABSA task.

{ We conducted additional experiments on various PLMs, including \(\text{BERT}_{\text{Large}}\), BART, and RoBERTa, as shown in Table \ref{tab:semeval_plm_acd}. From these experiments, we made the following observations.
	(1) BERT-ASC shows significant improvements over BERT-NLI in aspect detection. We attribute these improvements to the proficient generation of auxiliary sentences, which facilitates the model's ability to more effectively map the category to its corresponding representatives in the sentence.
	(2) However, these improvements are modest in aspect sentiment.
	(3) We also observe that BERT-ASC improves the stability measure in standard deviation, which is an important metric in multi-label classification scenarios.}

\subsubsection{Ablation Study}
{
	We conducted an ablation study to assess the contribution of each component of BERT-ASC to the final performance, as shown in Table \ref{tab:abl}. The analysis includes three versions of the model: BERT-ASC without semantic information (BERT-ASC W/O Semantic), BERT-ASC without syntactic information (BERT-ASC W/O Syntactic), and the full BERT-ASC model. The results demonstrate that the BERT-ASC W/O Semantic model contributes more to the final performance, as performance drops when it is eliminated. This is expected for the following reasons: syntactic information heavily relies on the accuracy of the parser, which may not always be reliable as shown in Table \ref{tab:error_analysis_1}. Sentences with implicit aspects (see Table \ref{tab:impliti_perc}) are left without opinion words, as we depend on the semantic latent space to generate the modifiers. The results show that the full BERT-ASC model outperforms its ablated versions across all metrics. Specifically, the BERT-ASC model achieves the highest Precision (96.17), Recall (92.04), and F1 score (94.05), indicating the significant contribution of both semantic and syntactic information to the model's performance.}

\begin{table*}[t]
	\centering

	\adjustbox{width=\linewidth}{
		\begin{tabular}{lcccccccc}
			\hline
			\makecell[c]{\multirow{2}{*}{Model}}&\multicolumn{3}{c}{Aspect Categorization}&& \multicolumn{2}{c}{Aspect Sentiment}\\\cline{2-4}\cline{6-8}
			&Strict Accuracy& Macro-F1&	AUC&&	Accuracy&	AUC&\\
			\hline
			LSTM-Loc \cite{saeidi2016sentihood}&  -& 69.3& 89.7&& 81.9& 83.9	\\				
			SenticLSTM \cite{ma2018targeted}&	 67.4& 78.2& -&& 89.3& -				\\
			Dmu-Entnet \cite{liu2018recurrent}&	 73.5& 78.5& 94.4&& 91.0& 94.8				\\
			
			BERT-single \cite{devlin2018bert}&	73.7&	81.0&	96.4&&	85.5&	84.2\\
			BERT-NLI-M \cite{sun2019utilizing}&78.3& 87.0& 97.5&& 92.1& 96.5	\\
			CG-BERT \cite{wu2020context}& 79.7 & 87.1 & 97.5&&93.7 &97.2\\
			BERT-NLI-B \cite{sun2019utilizing}&	79.8& 87.5& 96.6&&	92.8& 96.9\\
			QACG-BERT \cite{wu2020context}&	80.9 &	89.7&	97.8&&	93.8&	\textbf{97.8}\\
			\textbf{BERT-ASC (Ours)}&\textbf{85.17$\pm$0.58}&\textbf{94.0$\pm$0.38}&\textbf{99.09$\pm$0.06}&&\textbf{93.93$\pm$0.41}&{97.48$\pm$0.41}\\

			\hline
			
		\end{tabular}
	}
		\caption{Model performance on the SentiHood TABSA dataset, with the best performances highlighted in bold. Results are reported as the mean and standard deviation over 5 runs with different seeds. Comparative results are sourced from the original papers. The symbol ``-'' indicates that the result was not reported in the original paper.}
	\label{tab:TABSA_Results}
	
\end{table*}

\begin{table*}[t]
	\centering

	\adjustbox{width=\linewidth}{
		\begin{tabular}{llccccccccc}
			\hline
			\makecell[c]{\multirow{2}{*}{PLM}}&\makecell[c]{\multirow{2}{*}{Model}}&\multicolumn{3}{c}{Aspect Categorization}&& \multicolumn{2}{c}{Aspect Sentiment}\\\cline{3-5}\cline{7-9}
			&&Strict Accuracy& Macro-F1&	AUC&&	Accuracy&	AUC&\\
			\hline
			\multirow{2}{*}{\(\text{BART}_{\text{Base}}\)}&BERT-NLI&74.76$\pm$0.68&90.40$\pm$0.41&97.55$\pm$0.08&&94.45$\pm$0.14&97.74$\pm$0.19\\
			&BERT-ASC&84.46$\pm$1.05&94.34$\pm$0.12&99.20$\pm$0.09&&94.06$\pm$0.43&97.50$\pm$0.34\\
			\hline
			\multirow{2}{*}{\(\text{RoBERTa}_{\text{Base}}\)}&BERT-NLI&79.69$\pm$0.22&90.63$\pm$0.43&97.74$\pm$0.15&&95.00$\pm$0.42&97.66$\pm$0.18\\
			&BERT-ASC&86.55$\pm$0.54&94.50$\pm$0.26&99.18$\pm$0.05&&93.85$\pm$0.44&97.01$\pm$0.45\\
			\hline
			\multirow{2}{*}{\(\text{BART}_{\text{Large}}\)}&BERT-NLI&73.50$\pm$2.94&90.35$\pm$1.32&97.28$\pm$0.49&&94.57$\pm$0.97&97.85$\pm$0.62\\
			&BERT-ASC&85.50$\pm$1.53&94.68$\pm$0.57&99.21$\pm$0.14&&94.39$\pm$1.09&97.40$\pm$0.64\\
			\hline
			\multirow{2}{*}{\(\text{RoBERTa}_{\text{Large}}\)}&BERT-NLI&81.89$\pm$0.55&91.60$\pm$0.02&98.08$\pm$0.15&&95.29$\pm$0.44&98.20$\pm$0.30\\
			&BERT-ASC&85.50$\pm$1.53&94.68$\pm$0.57&99.21$\pm$0.14&&94.39$\pm$1.09&97.40$\pm$0.64\\
			
			\hline
			
			\multirow{2}{*}{\(\text{BERT}_{\text{Large}}\)}&BERT-NLI&79.19$\pm$0.77&89.52$\pm$0.41&97.34$\pm$0.19&&93.83$\pm$0.47&97.68$\pm$0.39\\
			&BERT-ASC&88.04$\pm$0.24&94.35$\pm$0.40&99.15$\pm$0.07&&94.16$\pm$0.20&97.20$\pm$0.37\\
			
			\hline

		\end{tabular}
	}
		\caption{{Evaluation results of SentiHood on various PLMs, averaged over five runs with different seeds, with standard deviations included. BERT-ASC demonstrates significant improvements in aspect detection and achieves competitive performance in sentiment detection. All results are based on our implementations.}}
	\label{tab:TABSA_Results_plm}
	
\end{table*}

\subsection{Exp-II: TABSA}

We validate the effectiveness of the proposed solution with the TABSA task on the SentiHood dataset. For a fair comparison, we follow the experimental settings of the comparative methods \cite{saeidi2016sentihood,ma2018targeted,sun2019utilizing,wu2020context}. Only four aspect categories (i.e., the most frequently { appearing} in the corpus) are considered (i.e., price, transit-location, safety, and general). For the aspect categorization subtask, we report the results of the strict accuracy metric, Macro-F1 score, and AUC. Strict accuracy requires the model to correctly detect all aspect categories for a given target, while Macro-F1 is the harmonic mean of the Macro-precision and Macro-recall of all targets. For the sentiment classification subtask, we report accuracy and AUC metrics.

In addition to the comparative methods of ABSA, including BERT-pair-NLI-B, BERT-pair-NLI-M, CG-BERT, and QACG-BERT \cite{sun2019utilizing,wu2020context}, we also compare our solution to the following models:

\begin{itemize}	
	\item \textbf{LSTM-Final} \cite{saeidi2016sentihood}. It used a bidirectional LSTM to learn a classifier for each category. It { regarded} the final state as the aspect-specific representation in response to the target;
	\item \textbf{LSTM-Location} \cite{saeidi2016sentihood}. It is a variant of LSTM-Final, which considered the representation of the target index as the aspect-specific representation{;}
	\item \textbf{SenticLSTM}  \cite{ma2018targeted}. An attentive biLSTM introduced, leveraging the external information as commonsense knowledge from SenticNet \cite{cambria2016senticnet}{;}
	\item \textbf{Dmu-Entnet} \cite{liu2018recurrent}. A bidirectional EntNet \cite{henaff2016tracking} that exploited the external memory chains with a delayed memory update mechanism to track entities{.}
	
\end{itemize}

\subsection{Results}

Consistent with Exp-I, we use the dev set to select the best model and average 5 run performances with different seeds, and report the detailed evaluation results { for} the SentiHood dataset in Table~\ref{tab:TABSA_Results} from which we have made the following observations. Compared to the traditional DNN-based approaches, our solution achieves the best performance with large margins in terms of strict accuracy, Macro-F1 { and} sentiment accuracy with 7.7\%, 12.6\% { and} 3\% improvements, respectively. Compared to the previous BERT-based state-of-the-art models, our solution consistently achieves state-of-the-art performance for aspect categorization and aspect-based sentiment subtasks in terms of strict Accuracy, Macro-F1 { and} sentiment accuracy with 81.2\%, 91.1\% { and} 94.0\%, respectively. Due to the widely recognized challenge of sentiment analysis, the achieved improvements can be deemed very considerable. This clearly suggests that a carefully-designed collaboration between prior knowledge and DNN can effectively perform better than a pure DNN on the TABSA task. {Table \ref{tab:TABSA_Results_plm} presents the performance results of various PLMs, including \(\text{BERT}_{\text{Large}}\), BART, and RoBERTa. Consistent with the SemEval, BERT-ASC demonstrates significant improvements over BERT-NLI in aspect detection. However, these improvements are relatively modest when it comes to aspect sentiment analysis}.

\begin{table*}
	\centering

	\adjustbox{width=\linewidth}{
		\begin{tabular}{lcccll}
			\hline
			\multirow{2}{*}{$s_i$}&\multirow{2}{*}{\makecell[c]{Sentence}} & \multirow{2}{*}{Annotated}&BERT-NLI&\multicolumn{2}{c}{BERT-ASC}\\\cline{5-6}
			& && \makecell[c]{Predicted}  & \makecell[c]{Predicted}& \makecell[c]{Auxiliary}\\
			\hline

			$s_1$&\makecell[l]{The service is excellent and always \\ informative without an air}&\makecell[c]{\{(S, POS)\}}&\makecell[c]{\{(S, POS), \\(A, POS)\}}&\makecell[c]{\{(S, POS)\}}&\makecell[c]{S=\{excellent,\\ informative\}}\\
			$s_2$&\makecell[l]{The wait staff was loud and inconsiderate}&\makecell[c]{\{(S, NEG)\}}&\makecell[c]{\{(S, NEG), \\(A, NEG)\}}&\makecell[c]{\{(S, NEG)\}}&\makecell[c]{S=\{staff, loud, \\inconsiderate\}}\\
			
			$s_3$ & \makecell[l]{The staff is unbelievably friendly\\ and I dream about their saag goshtso good} &\makecell[c]{\{(F, POS),\\ (S, POS)\}}&\makecell[c]{\{(F, POS)\}}&\makecell[c]{\{(F, POS), \\(S, POS)\}}&\makecell[c]{\ F=\{saag,good\},\\ S=\{staff,friendly\}}\\
			\hline
		\end{tabular}
	}
		\caption{A case study where the letters F, S, and A denote the aspects of food, service, and ambience, respectively, while POS and NEG represent positive and negative sentiments, respectively.}
	
	\label{tab:case_study}
\end{table*}
\begin{table*}
	\centering

	\adjustbox{width=\linewidth}{
		\begin{tabular}{llll}
			\hline
			$s_i$ &\makecell[c]{Sentence} &\makecell[c]{Annotated aspect}&Auxiliary-sentence\\
			\hline
			$s_1$ &The \textbf{atmosphere} was \textbf{nice} but it was a little too \textbf{dark}.&\{(ambience, conflict)\} &\{(atmosphere, nice)\}\\
			$s_2$ &I trust the people at go  sushi it never \textbf{disappoint}.&\{(anecdotes, positive)\} &\{(sushi, disappoint)\}\\
			$s_3$ &It shouldn't take \textbf{ten minutes} to get your drinks.& \{(service, negative)\}& \{(service,-)\}\\
			\hline
		\end{tabular}
	}
		\caption{An illustrative example of the error analysis.}
	\label{tab:error_analysis}
	
\end{table*}

\begin{table}
	\centering

		\begin{tabular}{lccc}
			\hline
			Dataset &Semantic& Syntax \\
			\hline
			SemEval&96.26 &58.2\%\\
			SentiHood&77.1\% &55.8\%\\
			\hline
		\end{tabular}
	
		\caption{{ The percentage of auxiliary-sentences that are automatically constructed.}}
	\label{tab:error_analysis_1}	
\end{table}

\subsection{Case study}

To better illustrate the effectiveness of BERT-ASC in terms of mapping implicit aspects to their indicators, we perform a case study shown in Table \ref{tab:case_study}. Particularly, we use the baseline BERT-NLI \cite{sun2019utilizing} as the comparative baseline. The reason behind this choice is that BERT-NLI employed the concept of sentence B in the setting of BERT \cite{devlin2018bert}, which { is} very related to BERT-ASC. It is noteworthy that QACG-BERT \cite{wu2020context} eliminated sentence B and introduced an attention mechanism to learn aspect-specific representation. Table \ref{tab:case_study} depicts some examples where BERT-NLI fails to accurately predict. As can be seen in \(s_1\), the presence of the word ``air'' leads BERT-NLI to wrongly associate the sentence with the aspect of ambience, while BERT-ASC does not construct an auxiliary-sentence for ambience.  
Similarly, BERT-NLI assigned an extra aspect of ambience to the sentence \(s_2\) due to the word ``loud'', which is often used to describe the environment. As mentioned in Section \ref{sec:intro}, mapping the implicit aspect to its indicators heavily relies on sufficient labeled examples, which may not be readily available in real scenarios. \(s_3\) is an illustrative example of such a scenario; the word ``saag'', which indicates food, may rarely occur in reviews and thus leads BERT-NLI to a wrong prediction. However, the collaboration between the food's seed and its semantic distribution in the embedding space encourages BERT-ASC to capture the token ``saag'' and its syntactic modifier ``good''.  All these examples suggest that a well-designed auxiliary sentence construction can significantly boost the performance of BERT in learning the implicit-aspect representation.

\subsection{Error Analysis}

For further improvement of our solution in the future, it is important to scrutinize the failure cases, which could be best classified into three major categories:

\begin{enumerate}
	\item \textbf{Inaccurate syntactic relations}. This type of error is { often} produced by the dependency parser, which is slightly tolerant to the informal expressions and complexity of online reviews \cite{li2021dual}. Consider the illustrative example in error analysis Table \ref{tab:error_analysis}, { where} the atmosphere in the sentence \(s_1\) is modified by { both} nice and dark, but only nice is detected by the parser and thus leads to { incorrect} polarity prediction.
	\item \textbf{Inaccurate semantic candidates}. This type of error { generally} occurs when a token is semantically close to the seed of an irrelevant aspect. For example, sushi in the sentence \(s_2\) is a candidate of { an} irrelevant aspect (i.e., food). 
	\item \textbf{Incomplete coverage}. This category of errors occurs when no semantic candidate is detected. For example, the service in sentence \(s_3\) is left without any candidates.
\end{enumerate}

\section{Conclusion}\label{sec:conclusion}

In this paper, we have proposed a novel solution to address { jointly} the implicit aspect categorization and sentiment subtasks in a unified framework. It first introduces a simple but effective mechanism to construct an auxiliary-sentence for the implicit aspect based on the semantic relatedness of the seed words of the aspect { targeted} in the embedding space. Then, it encourages BERT to learn the aspect-specific representation in response to the automatically generated auxiliary-sentence instead of the implicit aspect itself. Our extensive experiments have shown that the proposed approach consistently achieves state-of-the-art performance on both ABSA and TABSA tasks. For future work, it is interesting to note that incorporating prior knowledge with pre-trained language models, e.g., BERT, is potentially applicable to other classification tasks; however, technical solutions need further investigation.

\bibliographystyle{acl_natbib}
\bibliography{references}

\begin{thebibliography}{60}
\expandafter\ifx\csname natexlab\endcsname\relax\def\natexlab#1{#1}\fi

\bibitem[{Ahmed et~al.(2021)Ahmed, Chen, Wang, Nafa, Li, and
  Duan}]{ahmed-etal-2021-dnn}
Murtadha Ahmed, Qun Chen, Yanyan Wang, Youcef Nafa, Zhanhuai Li, and Tianyi
  Duan. 2021.
\newblock \href {https://doi.org/10.18653/v1/2021.findings-acl.43}
  {{{DNN}-driven Gradual Machine Learning for Aspect-term Sentiment Analysis}}.
\newblock In \emph{Proceedings of the 59th Annual Meeting of the Association
  for Computational Linguistics: {ACL/IJCNLP}, Findings}, pages 488--497.

\bibitem[{Ahmed et~al.(2023)Ahmed, Pan, Po, Su, Cao, Zhang, and
  Liu}]{ahmed2023rank}
Murtadha Ahmed, Shengfeng Pan, Wen Po, Jianlin Su, Xinxin Cao, Wenze Zhang, and
  Yunfeng Liu. 2023.
\newblock Rank-aware negative training for semi-supervised text classification.
\newblock \emph{Transactions of the Association for Computational Linguistics},
  11.

\bibitem[{Bickart and Schindler(2001)}]{bickart2001internet}
Barbara Bickart and Robert~M Schindler. 2001.
\newblock {Internet forums as influential sources of consumer information}.
\newblock \emph{Journal of interactive marketing}, 15(3):31--40.

\bibitem[{Bloom et~al.(2007)Bloom, Garg, and Argamon}]{bloom2007extracting}
Kenneth Bloom, Navendu Garg, and Shlomo Argamon. 2007.
\newblock \href {https://aclanthology.org/N07-1039/} {Extracting appraisal
  expressions}.
\newblock In \emph{Human Language Technology Conference of the North American
  Chapter of the Association of Computational Linguistics, Proceedings, April
  22-27, 2007, Rochester, New York, {USA}}, pages 308--315. The Association for
  Computational Linguistics.

\bibitem[{Brown et~al.(2020)Brown, Mann, Ryder, Subbiah, Kaplan, Dhariwal,
  Neelakantan, Shyam, Sastry, Askell, Agarwal, Herbert{-}Voss, Krueger,
  Henighan, Child, Ramesh, Ziegler, Wu, Winter, Hesse, Chen, Sigler, Litwin,
  Gray, Chess, Clark, Berner, McCandlish, Radford, Sutskever, and
  Amodei}]{BrownMRSKDNSSAA20}
Tom~B. Brown, Benjamin Mann, Nick Ryder, Melanie Subbiah, Jared Kaplan,
  Prafulla Dhariwal, Arvind Neelakantan, Pranav Shyam, Girish Sastry, Amanda
  Askell, Sandhini Agarwal, Ariel Herbert{-}Voss, Gretchen Krueger, Tom
  Henighan, Rewon Child, Aditya Ramesh, Daniel~M. Ziegler, Jeffrey Wu, Clemens
  Winter, Christopher Hesse, Mark Chen, Eric Sigler, Mateusz Litwin, Scott
  Gray, Benjamin Chess, Jack Clark, Christopher Berner, Sam McCandlish, Alec
  Radford, Ilya Sutskever, and Dario Amodei. 2020.
\newblock \href
  {https://proceedings.neurips.cc/paper/2020/hash/1457c0d6bfcb4967418bfb8ac142f64a-Abstract.html}
  {{Language Models are Few-Shot Learners}}.
\newblock In \emph{Advances in Neural Information Processing Systems 33: Annual
  Conference on Neural Information Processing Systems 2020, NeurIPS 2020,
  December 6-12, 2020, virtual}.

\bibitem[{Brun et~al.(2014)Brun, Popa, and Roux}]{brun2014xrce}
Caroline Brun, Diana~Nicoleta Popa, and Claude Roux. 2014.
\newblock \href {https://doi.org/10.3115/V1/S14-2149} {{XRCE:} hybrid
  classification for aspect-based sentiment analysis}.
\newblock In \emph{Proceedings of the 8th International Workshop on Semantic
  Evaluation, SemEval@COLING 2014, Dublin, Ireland, August 23-24, 2014}, pages
  838--842. The Association for Computer Linguistics.

\bibitem[{Cai et~al.(2021)Cai, Xia, and Yu}]{cai-etal-2021-aspect}
Hongjie Cai, Rui Xia, and Jianfei Yu. 2021.
\newblock \href {https://doi.org/10.18653/v1/2021.acl-long.29}
  {Aspect-category-opinion-sentiment quadruple extraction with implicit aspects
  and opinions}.
\newblock In \emph{Proceedings of the 59th Annual Meeting of the Association
  for Computational Linguistics and the 11th International Joint Conference on
  Natural Language Processing (Volume 1: Long Papers)}, pages 340--350, Online.
  Association for Computational Linguistics.

\bibitem[{Cambria et~al.(2016)Cambria, Poria, Bajpai, and
  Schuller}]{cambria2016senticnet}
Erik Cambria, Soujanya Poria, Rajiv Bajpai, and Bj{\"{o}}rn~W. Schuller. 2016.
\newblock \href {https://aclanthology.org/C16-1251/} {Senticnet 4: {A} semantic
  resource for sentiment analysis based on conceptual primitives}.
\newblock In \emph{{COLING} 2016, 26th International Conference on
  Computational Linguistics, Proceedings of the Conference: Technical Papers,
  December 11-16, 2016, Osaka, Japan}, pages 2666--2677. {ACL}.

\bibitem[{Castellucci et~al.(2014)Castellucci, Filice, Croce, and
  Basili}]{castellucci2014unitor}
Giuseppe Castellucci, Simone Filice, Danilo Croce, and Roberto Basili. 2014.
\newblock \href {https://doi.org/10.3115/V1/S14-2135} {{UNITOR:} aspect based
  sentiment analysis with structured learning}.
\newblock In \emph{Proceedings of the 8th International Workshop on Semantic
  Evaluation, SemEval@COLING 2014, Dublin, Ireland, August 23-24, 2014}, pages
  761--767. The Association for Computer Linguistics.

\bibitem[{Chen and Qian(2022)}]{ChenQ22}
Zhuang Chen and Tieyun Qian. 2022.
\newblock \href {https://doi.org/10.1109/TASLP.2022.3146052} {Retrieve-and-edit
  domain adaptation for end2end aspect based sentiment analysis}.
\newblock \emph{{IEEE} {ACM} Trans. Audio Speech Lang. Process.}, 30:659--672.

\bibitem[{Dai et~al.(2021)Dai, Yan, Sun, Liu, and Qiu}]{dai2021does}
Junqi Dai, Hang Yan, Tianxiang Sun, Pengfei Liu, and Xipeng Qiu. 2021.
\newblock \href {https://doi.org/10.18653/V1/2021.NAACL-MAIN.146} {Does syntax
  matter? {A} strong baseline for aspect-based sentiment analysis with
  roberta}.
\newblock In \emph{Proceedings of the 2021 Conference of the North American
  Chapter of the Association for Computational Linguistics: Human Language
  Technologies, {NAACL-HLT} 2021, Online, June 6-11, 2021}, pages 1816--1829.
  Association for Computational Linguistics.

\bibitem[{Davison et~al.(2019)Davison, Feldman, and
  Rush}]{davison2019commonsense}
Joe Davison, Joshua Feldman, and Alexander~M. Rush. 2019.
\newblock \href {https://doi.org/10.18653/V1/D19-1109} {Commonsense knowledge
  mining from pretrained models}.
\newblock In \emph{Proceedings of the 2019 Conference on Empirical Methods in
  Natural Language Processing and the 9th International Joint Conference on
  Natural Language Processing, {EMNLP-IJCNLP} 2019, Hong Kong, China, November
  3-7, 2019}, pages 1173--1178. Association for Computational Linguistics.

\bibitem[{de~Marneffe and Manning(2008)}]{de-marneffe-manning-2008-stanford}
Marie-Catherine de~Marneffe and Christopher~D. Manning. 2008.
\newblock \href {https://aclanthology.org/W08-1301} {The {S}tanford typed
  dependencies representation}.
\newblock In \emph{Coling 2008: Proceedings of the workshop on Cross-Framework
  and Cross-Domain Parser Evaluation}, pages 1--8, Manchester, UK. Coling 2008
  Organizing Committee.

\bibitem[{Devlin et~al.(2019)Devlin, Chang, Lee, and
  Toutanova}]{devlin2018bert}
Jacob Devlin, Ming{-}Wei Chang, Kenton Lee, and Kristina Toutanova. 2019.
\newblock \href {https://doi.org/10.18653/v1/n19-1423} {{{BERT:} Pre-training
  of Deep Bidirectional Transformers for Language Understanding}}.
\newblock In \emph{Proceedings of the 2019 Conference of the North American
  Chapter of the Association for Computational Linguistics: Human Language
  Technologies, {NAACL-HLT}}, pages 4171--4186.

\bibitem[{Du et~al.(2020)Du, Sun, Wang, Qi, and Liao}]{du2020adversarial}
Chunning Du, Haifeng Sun, Jingyu Wang, Qi~Qi, and Jianxin Liao. 2020.
\newblock \href {https://doi.org/10.18653/V1/2020.ACL-MAIN.370} {Adversarial
  and domain-aware {BERT} for cross-domain sentiment analysis}.
\newblock In \emph{Proceedings of the 58th Annual Meeting of the Association
  for Computational Linguistics, {ACL} 2020, Online, July 5-10, 2020}, pages
  4019--4028. Association for Computational Linguistics.

\bibitem[{Han et~al.(2021)Han, Zhang, Ding, Gu, Liu, Huo, Qiu, Yao, Zhang,
  Zhang, Han, Huang, Jin, Lan, Liu, Liu, Lu, Qiu, Song, Tang, Wen, Yuan, Zhao,
  and Zhu}]{HanZDGLHQYZZHHJ21}
Xu~Han, Zhengyan Zhang, Ning Ding, Yuxian Gu, Xiao Liu, Yuqi Huo, Jiezhong Qiu,
  Yuan Yao, Ao~Zhang, Liang Zhang, Wentao Han, Minlie Huang, Qin Jin, Yanyan
  Lan, Yang Liu, Zhiyuan Liu, Zhiwu Lu, Xipeng Qiu, Ruihua Song, Jie Tang,
  Ji{-}Rong Wen, Jinhui Yuan, Wayne~Xin Zhao, and Jun Zhu. 2021.
\newblock \href {https://doi.org/10.1016/j.aiopen.2021.08.002} {{Pre-trained
  models: Past, present and future}}.
\newblock \emph{{AI} Open}, 2:225--250.

\bibitem[{Hao et~al.(2022)Hao, Sun, Dong, Han, Gu, and Wei}]{Yaru-2212-06713}
Yaru Hao, Yutao Sun, Li~Dong, Zhixiong Han, Yuxian Gu, and Furu Wei. 2022.
\newblock \href {https://doi.org/10.48550/arXiv.2212.06713} {{Structured
  Prompting: Scaling In-Context Learning to 1,000 Examples}}.
\newblock \emph{CoRR}, abs/2212.06713.

\bibitem[{He et~al.(2017)He, Lee, Ng, and Dahlmeier}]{he2017unsupervised}
Ruidan He, Wee~Sun Lee, Hwee~Tou Ng, and Daniel Dahlmeier. 2017.
\newblock \href {https://doi.org/10.18653/V1/P17-1036} {An unsupervised neural
  attention model for aspect extraction}.
\newblock In \emph{Proceedings of the 55th Annual Meeting of the Association
  for Computational Linguistics, {ACL} 2017, Vancouver, Canada, July 30 -
  August 4, Volume 1: Long Papers}, pages 388--397. Association for
  Computational Linguistics.

\bibitem[{Henaff et~al.(2017)Henaff, Weston, Szlam, Bordes, and
  LeCun}]{henaff2016tracking}
Mikael Henaff, Jason Weston, Arthur Szlam, Antoine Bordes, and Yann LeCun.
  2017.
\newblock \href {https://openreview.net/forum?id=rJTKKKqeg} {Tracking the world
  state with recurrent entity networks}.
\newblock In \emph{5th International Conference on Learning Representations,
  {ICLR} 2017, Toulon, France, April 24-26, 2017, Conference Track
  Proceedings}. OpenReview.net.

\bibitem[{Hu and Liu(2004)}]{hu2004mining}
Minqing Hu and Bing Liu. 2004.
\newblock \href {https://doi.org/10.1145/1014052.1014073} {{Mining and
  summarizing customer reviews}}.
\newblock In \emph{Proceedings of the Tenth {ACM} {SIGKDD} International
  Conference on Knowledge Discovery and Data Mining}, pages 168--177.

\bibitem[{Jawahar et~al.(2019)Jawahar, Sagot, and
  Seddah}]{jawahar-etal-2019-BERT}
Ganesh Jawahar, Beno{\^\i}t Sagot, and Djam{\'e} Seddah. 2019.
\newblock \href {https://doi.org/10.18653/v1/P19-1356} {What does {BERT} learn
  about the structure of language?}
\newblock In \emph{Proceedings of the 57th Annual Meeting of the Association
  for Computational Linguistics}, pages 3651--3657, Florence, Italy.
  Association for Computational Linguistics.

\bibitem[{Karimi et~al.(2021)Karimi, Rossi, and Prati}]{karimi2020improving}
Akbar Karimi, Leonardo Rossi, and Andrea Prati. 2021.
\newblock \href {https://aclanthology.org/2021.icnlsp-1.5} {Improving {BERT}
  performance for aspect-based sentiment analysis}.
\newblock In \emph{4th International Conference on Natural Language and Speech
  Processing, Trento, Italy, November 12-13, 2021}, pages 196--203. Association
  for Computational Linguistics.

\bibitem[{Kiritchenko et~al.(2014)Kiritchenko, Zhu, Cherry, and
  Mohammad}]{kiritchenko2014nrc}
Svetlana Kiritchenko, Xiaodan Zhu, Colin Cherry, and Saif~M. Mohammad. 2014.
\newblock \href {https://doi.org/10.3115/V1/S14-2076} {{NRC-Canada-2014:
  Detecting Aspects and Sentiment in Customer Reviews}}.
\newblock In \emph{Proceedings of the 8th International Workshop on Semantic
  Evaluation, SemEval@COLING 2014, Dublin, Ireland, August 23-24, 2014}, pages
  437--442. The Association for Computer Linguistics.

\bibitem[{Li et~al.(2022)Li, Zhao, Jin, Li, Shen, Tao, and Tao}]{LiZJLSTT22}
Jia Li, Yuyuan Zhao, Zhi Jin, Ge~Li, Tao Shen, Zhengwei Tao, and Chongyang Tao.
  2022.
\newblock \href {https://doi.org/10.1145/3511808.3557452} {{SK2:} integrating
  implicit sentiment knowledge and explicit syntax knowledge for aspect-based
  sentiment analysis}.
\newblock In \emph{Proceedings of the 31st {ACM} International Conference on
  Information {\&} Knowledge Management, Atlanta, GA, USA, October 17-21,
  2022}, pages 1114--1123. {ACM}.

\bibitem[{Li et~al.(2021)Li, Chen, Feng, Ma, Wang, and Hovy}]{li2021dual}
Ruifan Li, Hao Chen, Fangxiang Feng, Zhanyu Ma, Xiaojie Wang, and Eduard~H.
  Hovy. 2021.
\newblock \href {https://doi.org/10.18653/V1/2021.ACL-LONG.494} {Dual graph
  convolutional networks for aspect-based sentiment analysis}.
\newblock In \emph{Proceedings of the 59th Annual Meeting of the Association
  for Computational Linguistics and the 11th International Joint Conference on
  Natural Language Processing, {ACL/IJCNLP} 2021, (Volume 1: Long Papers),
  Virtual Event, August 1-6, 2021}, pages 6319--6329. Association for
  Computational Linguistics.

\bibitem[{Lin et~al.(2017)Lin, Doll{\'{a}}r, Girshick, He, Hariharan, and
  Belongie}]{lin2017feature}
Tsung{-}Yi Lin, Piotr Doll{\'{a}}r, Ross~B. Girshick, Kaiming He, Bharath
  Hariharan, and Serge~J. Belongie. 2017.
\newblock \href {https://doi.org/10.1109/CVPR.2017.106} {Feature pyramid
  networks for object detection}.
\newblock In \emph{2017 {IEEE} Conference on Computer Vision and Pattern
  Recognition, {CVPR} 2017, Honolulu, HI, USA, July 21-26, 2017}, pages
  936--944. {IEEE} Computer Society.

\bibitem[{Liu(2012)}]{liu2012sentiment}
Bing Liu. 2012.
\newblock Sentiment analysis and opinion mining.
\newblock \emph{Synthesis lectures on human language technologies},
  5(1):1--167.

\bibitem[{Liu et~al.(2018)Liu, Cohn, and Baldwin}]{liu2018recurrent}
Fei Liu, Trevor Cohn, and Timothy Baldwin. 2018.
\newblock \href {https://doi.org/10.18653/V1/N18-2045} {Recurrent entity
  networks with delayed memory update for targeted aspect-based sentiment
  analysis}.
\newblock In \emph{Proceedings of the 2018 Conference of the North American
  Chapter of the Association for Computational Linguistics: Human Language
  Technologies, NAACL-HLT, New Orleans, Louisiana, USA, June 1-6, 2018, Volume
  2 (Short Papers)}, pages 278--283. Association for Computational Linguistics.

\bibitem[{Liu et~al.(2019)Liu, Ott, Goyal, Du, Joshi, Chen, Levy, Lewis,
  Zettlemoyer, and Stoyanov}]{abs-1907-11692}
Yinhan Liu, Myle Ott, Naman Goyal, Jingfei Du, Mandar Joshi, Danqi Chen, Omer
  Levy, Mike Lewis, Luke Zettlemoyer, and Veselin Stoyanov. 2019.
\newblock \href {http://arxiv.org/abs/1907.11692} {Roberta: {A} robustly
  optimized {BERT} pretraining approach}.
\newblock \emph{CoRR}, abs/1907.11692.

\bibitem[{Ma et~al.(2018)Ma, Peng, and Cambria}]{ma2018targeted}
Yukun Ma, Haiyun Peng, and Erik Cambria. 2018.
\newblock \href {https://doi.org/10.1609/AAAI.V32I1.12048} {Targeted
  aspect-based sentiment analysis via embedding commonsense knowledge into an
  attentive {LSTM}}.
\newblock In \emph{Proceedings of the Thirty-Second {AAAI} Conference on
  Artificial Intelligence, (AAAI-18), the 30th innovative Applications of
  Artificial Intelligence (IAAI-18), and the 8th {AAAI} Symposium on
  Educational Advances in Artificial Intelligence (EAAI-18), New Orleans,
  Louisiana, USA, February 2-7, 2018}, pages 5876--5883. {AAAI} Press.

\bibitem[{Mikolov et~al.(2013)Mikolov, Yih, and Zweig}]{mikolov2013linguistic}
Tom{\'{a}}s Mikolov, Wen{-}tau Yih, and Geoffrey Zweig. 2013.
\newblock \href {https://aclanthology.org/N13-1090/} {Linguistic regularities
  in continuous space word representations}.
\newblock In \emph{Human Language Technologies: Conference of the North
  American Chapter of the Association of Computational Linguistics,
  Proceedings, June 9-14, 2013, Westin Peachtree Plaza Hotel, Atlanta, Georgia,
  {USA}}, pages 746--751. The Association for Computational Linguistics.

\bibitem[{Movahedi et~al.(2019)Movahedi, Ghadery, Faili, and
  Shakery}]{movahedi2019aspect}
Sajad Movahedi, Erfan Ghadery, Heshaam Faili, and Azadeh Shakery. 2019.
\newblock \href {http://arxiv.org/abs/1901.01183} {Aspect category detection
  via topic-attention network}.
\newblock \emph{CoRR}, abs/1901.01183.

\bibitem[{Murtadha et~al.(2020)Murtadha, Chen, and Li}]{ahmed2020constructing}
Ahmed Murtadha, Qun Chen, and Zhanhuai Li. 2020.
\newblock \href {https://doi.org/10.1007/s00521-020-04824-8} {{Constructing
  Domain-Dependent sentiment Dictionary for Sentiment Analysis}}.
\newblock \emph{Neural Comput. Appl.}, 32(18):14719--14732.

\bibitem[{Peters et~al.(2019)Peters, Ruder, and Smith}]{peters2019tune}
Matthew~E. Peters, Sebastian Ruder, and Noah~A. Smith. 2019.
\newblock \href {https://doi.org/10.18653/V1/W19-4302} {To tune or not to tune?
  adapting pretrained representations to diverse tasks}.
\newblock In \emph{Proceedings of the 4th Workshop on Representation Learning
  for NLP, RepL4NLP@ACL 2019, Florence, Italy, August 2, 2019}, pages 7--14.
  Association for Computational Linguistics.

\bibitem[{Pontiki et~al.(2014)Pontiki, Galanis, Pavlopoulos, Papageorgiou,
  Androutsopoulos, and Manandhar}]{pontiki2014semeval}
Maria Pontiki, Dimitris Galanis, John Pavlopoulos, Harris Papageorgiou, Ion
  Androutsopoulos, and Suresh Manandhar. 2014.
\newblock \href {https://doi.org/10.3115/v1/S14-2004} {{S}em{E}val-2014 task 4:
  Aspect based sentiment analysis}.
\newblock In \emph{Proceedings of the 8th International Workshop on Semantic
  Evaluation ({S}em{E}val 2014)}, pages 27--35, Dublin, Ireland. Association
  for Computational Linguistics.

\bibitem[{Popescu and Etzioni(2005)}]{popescu2007extracting}
Ana{-}Maria Popescu and Oren Etzioni. 2005.
\newblock \href {https://aclanthology.org/H05-1043/} {Extracting product
  features and opinions from reviews}.
\newblock In \emph{{HLT/EMNLP} 2005, Human Language Technology Conference and
  Conference on Empirical Methods in Natural Language Processing, Proceedings
  of the Conference, 6-8 October 2005, Vancouver, British Columbia, Canada},
  pages 339--346. The Association for Computational Linguistics.

\bibitem[{Qiu et~al.(2011)Qiu, Liu, Bu, and Chen}]{qiu2011opinion}
Guang Qiu, Bing Liu, Jiajun Bu, and Chun Chen. 2011.
\newblock \href {https://doi.org/10.1162/COLI\_A\_00034} {Opinion word
  expansion and target extraction through double propagation}.
\newblock \emph{Comput. Linguistics}, 37(1):9--27.

\bibitem[{Qiu et~al.(2020)Qiu, Sun, Xu, Shao, Dai, and Huang}]{Qiu-2003-08271}
Xipeng Qiu, Tianxiang Sun, Yige Xu, Yunfan Shao, Ning Dai, and Xuanjing Huang.
  2020.
\newblock \href {http://arxiv.org/abs/2003.08271} {{Pre-trained Models for
  Natural Language Processing: {A} Survey}}.
\newblock \emph{CoRR}, abs/2003.08271.

\bibitem[{Ramage et~al.(2009{\natexlab{a}})Ramage, Hall, Nallapati, and
  Manning}]{ramage-etal-2009-labeled}
Daniel Ramage, David Hall, Ramesh Nallapati, and Christopher~D. Manning.
  2009{\natexlab{a}}.
\newblock \href {https://aclanthology.org/D09-1026} {{L}abeled {LDA}: A
  supervised topic model for credit attribution in multi-labeled corpora}.
\newblock In \emph{Proceedings of the 2009 Conference on Empirical Methods in
  Natural Language Processing}, pages 248--256, Singapore. Association for
  Computational Linguistics.

\bibitem[{Ramage et~al.(2009{\natexlab{b}})Ramage, Hall, Nallapati, and
  Manning}]{ramage2009labeled}
Daniel Ramage, David Hall, Ramesh Nallapati, and Christopher~D Manning.
  2009{\natexlab{b}}.
\newblock Labeled lda: A supervised topic model for credit attribution in
  multi-labeled corpora.
\newblock In \emph{Proceedings of the 2009 conference on empirical methods in
  natural language processing}, pages 248--256.

\bibitem[{Ratner et~al.(2023)Ratner, Levine, Belinkov, Ram, Magar, Abend,
  Karpas, Shashua, Leyton{-}Brown, and Shoham}]{pcw}
Nir Ratner, Yoav Levine, Yonatan Belinkov, Ori Ram, Inbal Magar, Omri Abend,
  Ehud Karpas, Amnon Shashua, Kevin Leyton{-}Brown, and Yoav Shoham. 2023.
\newblock \href {https://doi.org/10.18653/v1/2023.acl-long.352} {Parallel
  context windows for large language models}.
\newblock In \emph{Proceedings of the 61st Annual Meeting of the Association
  for Computational Linguistics (Volume 1: Long Papers), {ACL} 2023, Toronto,
  Canada, July 9-14, 2023}, pages 6383--6402. Association for Computational
  Linguistics.

\bibitem[{Reimers and Gurevych(2019)}]{reimers2019sentence}
Nils Reimers and Iryna Gurevych. 2019.
\newblock \href {https://doi.org/10.18653/V1/D19-1410} {Sentence-bert: Sentence
  embeddings using siamese bert-networks}.
\newblock In \emph{Proceedings of the 2019 Conference on Empirical Methods in
  Natural Language Processing and the 9th International Joint Conference on
  Natural Language Processing, {EMNLP-IJCNLP} 2019, Hong Kong, China, November
  3-7, 2019}, pages 3980--3990. Association for Computational Linguistics.

\bibitem[{Saeidi et~al.(2016{\natexlab{a}})Saeidi, Bouchard, Liakata, and
  Riedel}]{saeidi-etal-2016-sentihood}
Marzieh Saeidi, Guillaume Bouchard, Maria Liakata, and Sebastian Riedel.
  2016{\natexlab{a}}.
\newblock \href {https://www.aclweb.org/anthology/C16-1146} {{S}enti{H}ood:
  Targeted aspect based sentiment analysis dataset for urban neighbourhoods}.
\newblock In \emph{Proceedings of {COLING} 2016, the 26th International
  Conference on Computational Linguistics: Technical Papers}, pages 1546--1556,
  Osaka, Japan. The COLING 2016 Organizing Committee.

\bibitem[{Saeidi et~al.(2016{\natexlab{b}})Saeidi, Bouchard, Liakata, and
  Riedel}]{saeidi2016sentihood}
Marzieh Saeidi, Guillaume Bouchard, Maria Liakata, and Sebastian Riedel.
  2016{\natexlab{b}}.
\newblock \href {https://aclanthology.org/C16-1146/} {Sentihood: Targeted
  aspect based sentiment analysis dataset for urban neighbourhoods}.
\newblock In \emph{{COLING} 2016, 26th International Conference on
  Computational Linguistics, Proceedings of the Conference: Technical Papers,
  December 11-16, 2016, Osaka, Japan}, pages 1546--1556. {ACL}.

\bibitem[{Schouten and Frasincar(2016)}]{Schouten2016tkde}
Kim Schouten and Flavius Frasincar. 2016.
\newblock \href {https://doi.org/10.1109/TKDE.2015.2485209} {{Survey on
  Aspect-Level Sentiment Analysis}}.
\newblock \emph{{IEEE} Transactions on Knowledge and Data Engineering {TKDE}},
  28(3):813--830.

\bibitem[{Schouten et~al.(2018)Schouten, van~der Weijde, Frasincar, and
  Dekker}]{schouten2017supervised}
Kim Schouten, Onne van~der Weijde, Flavius Frasincar, and Rommert Dekker. 2018.
\newblock \href {https://doi.org/10.1109/TCYB.2017.2688801} {Supervised and
  unsupervised aspect category detection for sentiment analysis with
  co-occurrence data}.
\newblock \emph{{IEEE} Trans. Cybern.}, 48(4):1263--1275.

\bibitem[{Su et~al.(2024)Su, Ahmed, Bo, Ao, Zhu, and Liu}]{nbce}
Jianlin Su, Murtadha Ahmed, Wen Bo, Luo Ao, Mingren Zhu, and Yunfeng Liu. 2024.
\newblock \href {https://doi.org/10.48550/ARXIV.2403.17552} {Naive bayes-based
  context extension for large language models}.
\newblock \emph{CoRR}, abs/2403.17552.

\bibitem[{Sun et~al.(2019)Sun, Huang, and Qiu}]{sun2019utilizing}
Chi Sun, Luyao Huang, and Xipeng Qiu. 2019.
\newblock \href {https://doi.org/10.18653/V1/N19-1035} {Utilizing {BERT} for
  aspect-based sentiment analysis via constructing auxiliary sentence}.
\newblock In \emph{Proceedings of the 2019 Conference of the North American
  Chapter of the Association for Computational Linguistics: Human Language
  Technologies, {NAACL-HLT} 2019, Minneapolis, MN, USA, June 2-7, 2019, Volume
  1 (Long and Short Papers)}, pages 380--385. Association for Computational
  Linguistics.

\bibitem[{Toh and Su(2016)}]{toh2016nlangp}
Zhiqiang Toh and Jian Su. 2016.
\newblock \href {https://doi.org/10.18653/V1/S16-1045} {{NLANGP} at
  semeval-2016 task 5: Improving aspect based sentiment analysis using neural
  network features}.
\newblock In \emph{Proceedings of the 10th International Workshop on Semantic
  Evaluation, SemEval@NAACL-HLT 2016, San Diego, CA, USA, June 16-17, 2016},
  pages 282--288. The Association for Computer Linguistics.

\bibitem[{Venugopalan and Gupta(2022)}]{VenugopalanG22}
Manju Venugopalan and Deepa Gupta. 2022.
\newblock \href {https://doi.org/10.1016/J.KNOSYS.2022.108668} {An enhanced
  guided {LDA} model augmented with {BERT} based semantic strength for aspect
  term extraction in sentiment analysis}.
\newblock \emph{Knowl. Based Syst.}, 246:108668.

\bibitem[{Wan et~al.(2020)Wan, Yang, Du, Liu, Qi, and Pan}]{wan2020target}
Hai Wan, Yufei Yang, Jianfeng Du, Yanan Liu, Kunxun Qi, and Jeff~Z. Pan. 2020.
\newblock \href {https://doi.org/10.1609/AAAI.V34I05.6447}
  {Target-aspect-sentiment joint detection for aspect-based sentiment
  analysis}.
\newblock In \emph{The Thirty-Fourth {AAAI} Conference on Artificial
  Intelligence, {AAAI} 2020, The Thirty-Second Innovative Applications of
  Artificial Intelligence Conference, {IAAI} 2020, The Tenth {AAAI} Symposium
  on Educational Advances in Artificial Intelligence, {EAAI} 2020, New York,
  NY, USA, February 7-12, 2020}, pages 9122--9129. {AAAI} Press.

\bibitem[{Wang et~al.(2022)Wang, Yang, Huang, Jiao, Yang, Jiang, Majumder, and
  Wei}]{wang2022simlm}
Liang Wang, Nan Yang, Xiaolong Huang, Binxing Jiao, Linjun Yang, Daxin Jiang,
  Rangan Majumder, and Furu Wei. 2022.
\newblock \href {https://doi.org/10.48550/ARXIV.2212.03533} {Text embeddings by
  weakly-supervised contrastive pre-training}.
\newblock \emph{CoRR}, abs/2212.03533.

\bibitem[{Wu et~al.(2021)Wu, Xiong, Yi, Yu, Zhu, Gao, and Chen}]{WU2021107073}
Chao Wu, Qingyu Xiong, Hualing Yi, Yang Yu, Qiwu Zhu, Min Gao, and Jie Chen.
  2021.
\newblock \href {https://doi.org/https://doi.org/10.1016/j.knosys.2021.107073}
  {Multiple-element joint detection for aspect-based sentiment analysis}.
\newblock \emph{Knowledge-Based Systems}, 223:107073.

\bibitem[{Wu and Ong(2021)}]{wu2020context}
Zhengxuan Wu and Desmond~C. Ong. 2021.
\newblock \href {https://doi.org/10.1609/AAAI.V35I16.17659} {Context-guided
  {BERT} for targeted aspect-based sentiment analysis}.
\newblock In \emph{Thirty-Fifth {AAAI} Conference on Artificial Intelligence,
  {AAAI} 2021, Thirty-Third Conference on Innovative Applications of Artificial
  Intelligence, {IAAI} 2021, The Eleventh Symposium on Educational Advances in
  Artificial Intelligence, {EAAI} 2021, Virtual Event, February 2-9, 2021},
  pages 14094--14102. {AAAI} Press.

\bibitem[{Xu et~al.(2019)Xu, Liu, Shu, and Yu}]{xu2019BERT}
Hu~Xu, Bing Liu, Lei Shu, and Philip~S. Yu. 2019.
\newblock \href {https://doi.org/10.18653/V1/N19-1242} {{BERT} post-training
  for review reading comprehension and aspect-based sentiment analysis}.
\newblock In \emph{Proceedings of the 2019 Conference of the North American
  Chapter of the Association for Computational Linguistics: Human Language
  Technologies, {NAACL-HLT} 2019, Minneapolis, MN, USA, June 2-7, 2019, Volume
  1 (Long and Short Papers)}, pages 2324--2335. Association for Computational
  Linguistics.

\bibitem[{Xue et~al.(2017)Xue, Zhou, Li, and Wang}]{xue2017mtna}
Wei Xue, Wubai Zhou, Tao Li, and Qing Wang. 2017.
\newblock \href {https://aclanthology.org/I17-2026/} {{MTNA:} {A} neural
  multi-task model for aspect category classification and aspect term
  extraction on restaurant reviews}.
\newblock In \emph{Proceedings of the Eighth International Joint Conference on
  Natural Language Processing, {IJCNLP} 2017, Taipei, Taiwan, November 27 -
  December 1, 2017, Volume 2: Short Papers}, pages 151--156. Asian Federation
  of Natural Language Processing.

\bibitem[{Zeng et~al.(2019)Zeng, Zhou, Liu, and Song}]{zeng2019variational}
Ziqian Zeng, Wenxuan Zhou, Xin Liu, and Yangqiu Song. 2019.
\newblock \href {https://doi.org/10.18653/V1/N19-1036} {A variational approach
  to weakly supervised document-level multi-aspect sentiment classification}.
\newblock In \emph{Proceedings of the 2019 Conference of the North American
  Chapter of the Association for Computational Linguistics: Human Language
  Technologies, {NAACL-HLT} 2019, Minneapolis, MN, USA, June 2-7, 2019, Volume
  1 (Long and Short Papers)}, pages 386--396. Association for Computational
  Linguistics.

\bibitem[{Zhao et~al.(2022)Zhao, Tang, Tang, Wang, and Chen}]{ZhaoTTWC22}
Ziguo Zhao, Mingwei Tang, Wei Tang, Chunhao Wang, and Xiaoliang Chen. 2022.
\newblock \href {https://doi.org/10.1016/J.NEUCOM.2022.05.045} {Graph
  convolutional network with multiple weight mechanisms for aspect-based
  sentiment analysis}.
\newblock \emph{Neurocomputing}, 500:124--134.

\bibitem[{Zhou et~al.(2023)Zhou, Shen, Chen, and Cao}]{Zhou0CC23}
Ting Zhou, Ying Shen, Kang Chen, and Qing Cao. 2023.
\newblock \href {https://doi.org/10.1016/J.KNOSYS.2023.110740} {Hierarchical
  dual graph convolutional network for aspect-based sentiment analysis}.
\newblock \emph{Knowl. Based Syst.}, 276:110740.

\bibitem[{Zhou et~al.(2015)Zhou, Wan, and Xiao}]{zhou2015representation}
Xinjie Zhou, Xiaojun Wan, and Jianguo Xiao. 2015.
\newblock \href {https://doi.org/10.1609/AAAI.V29I1.9194} {Representation
  learning for aspect category detection in online reviews}.
\newblock In \emph{Proceedings of the Twenty-Ninth {AAAI} Conference on
  Artificial Intelligence, January 25-30, 2015, Austin, Texas, {USA}}, pages
  417--424. {AAAI} Press.

\end{thebibliography}

\end{document}